\documentclass[11pt,a4paper]{article}
\usepackage{memoraxpaper}
\usepackage{algorithm}
\usepackage{algpseudocode}
\usepackage[
  backend=biber,
  style=authoryear-comp,
  maxcitenames=2,
  maxbibnames=12,
  uniquename=init,
  giveninits=true,
  doi=true,
  url=true,
  isbn=false,
  eprint=true
]{biblatex}
\AtEveryCite{\color{MemoraXBlue}}
\AtBeginBibliography{\raggedright\sloppy}

\reporttitle{From Memory to Guide: Spatio-Temporal Composer for Procedural Coding Memory}
\reportrunningtitle{Composer}
\reportauthors{Zhixuan Tan\textsuperscript{1,*}, Pengjie Gu\textsuperscript{2,*}, Zhao Li\textsuperscript{2}, Yihan Hu\textsuperscript{2}, Xu He\textsuperscript{2}, Dong Li\textsuperscript{2}, Jianye Hao\textsuperscript{2}}
\reportaffiliations{\textsuperscript{1}The Chinese University of Hong Kong, Shenzhen\\\textsuperscript{2}MemoraX AI}
\reportversion{V 1.0}
\reportdate{Sep 2026}
\reportcontact{%
  \textsuperscript{*}Equal contribution.\\
  Corresponding author:
  \href{mailto:jianye.hao@tju.edu.cn}{jianye.hao@tju.edu.cn}%
}
\reportfooter{MemoraX AI Research · Composer}
\reportlogo{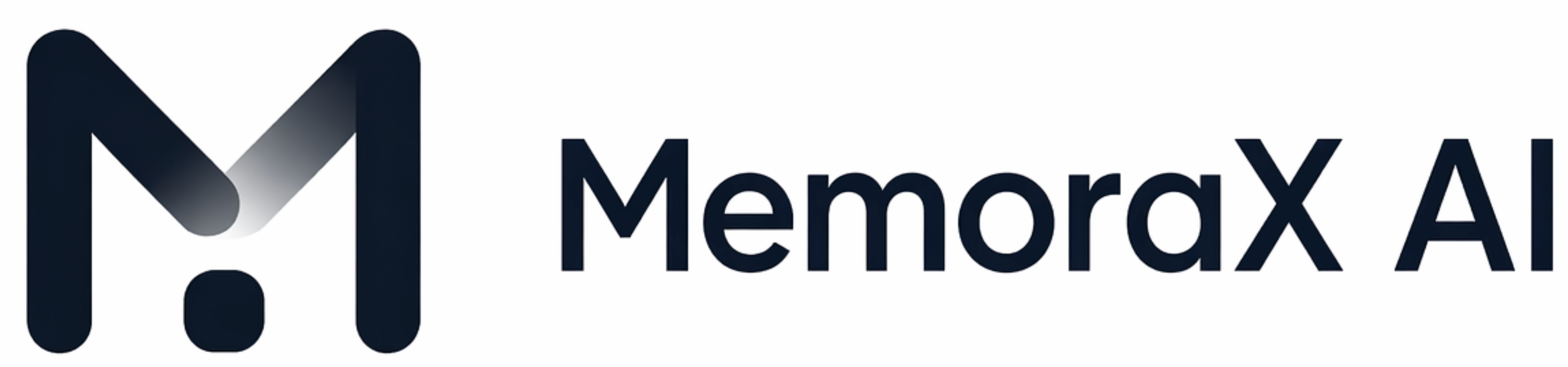}
\reportkeywords{agent memory, skill evolution, recursive self-improvement}

\begin{document}

\makememoraxpapertitle

\begin{reportabstract}
Memory-augmented agents typically integrate procedural knowledge by injecting retrieved skills directly into text prompts. This approach dangerously equates readable text with reliable execution. To bridge this gap, we introduce \textit{From Memory to Guide}, a novel paradigm that transitions procedural memory from passive text delivery to active, inference-time policy adaptation. We instantiate this paradigm through the Spatio-Temporal Composer, an active policy compiler that explicitly manages exactly how and when retrieved knowledge should be applied. Rather than treating skills as plug-and-play modules, Composer dynamically aligns historical knowledge with current environmental constraints (\emph{spatial adaptation}) and precisely dictates its applicable lifecycle (\emph{temporal orchestration}). It actively transforms static memories into strictly bounded \textit{Runtime Guides}---equipping the agent with localized objectives and behavioral guardrails without requiring a single parameter update. Extensive evaluations on 13 demanding, long-horizon software engineering tasks in EngramBench demonstrate the clear advantages of this architecture. Composer not only robustly prevents context mismatch but drives an absolute pass-rate increase of 7.2 percentage points on the most complex tasks, while simultaneously slashing the main agent's token usage by 32.2\%.
\end{reportabstract}

\printreportkeywords

\section{Introduction}
\label{sec:introduction}

LLMs are rapidly evolving from passive chatbots into autonomous agents executing complex real-world workflows. A fundamental bottleneck in this shift is the gap between general knowledge and practical execution. While LLMs possess vast amounts of pre-trained knowledge, their static weights struggle to capture the dynamic, hands-on experience required for lengthy tasks like software engineering. To bridge this gap, \textit{procedural memory} has emerged as a critical infrastructure. By enabling agents to accumulate step-by-step procedures, decision rules, and debugging experience as reusable \textit{Skills}, procedural memory empowers agents to continuously learn and reuse know-how without the prohibitive costs of retraining. Consequently, equipping agents with reusable skills has become a standard practice, driving recent successes in autonomous execution (e.g., ReMe~\parencite{ReMe}, Mem$^p$~\parencite{memp_2026}, and SkillRL~\parencite{skillrl_2026}).

However, simply accumulating experience exposes a critical vulnerability: the illusion of direct applicability. In long-horizon environments, directly injecting retrieved skills without contextual adaptation is not merely suboptimal—it is frequently catastrophic. When a retrieved skill encodes assumptions from a mismatched source context, blindly executing this static guidance introduces severe regressions. Consider a \emph{spatial} mismatch: if a static Skill explicitly dictates using \texttt{npm} commands, but the target project strictly enforces \texttt{yarn}, a direct application will instantly corrupt the lockfile and crash the entire build system. Similarly, consider a \emph{temporal} mismatch: because software engineering tasks are highly sequential, executing the right Skill at the wrong time triggers severe cascading consequences. For example, if an architectural Skill is triggered during a simple bug fix, the agent might rewrite working code instead of just patching the error. Consider a Skill that correctly guides architectural refactoring during initial scaffolding. If this exact same Skill is retrieved during late-stage debugging, it becomes actively destructive—prompting the agent to dismantle working code rather than fixing a simple localized bug. These spatial and temporal mismatches completely derail the agent's reasoning trajectory, triggering cascading errors and trapping it in irreversible debugging loops. 

Why do current memory-augmented agents struggle to generalize? State-of-the-art frameworks—like \textbf{AtomMem}~\parencite{atommem_2026}, \textbf{MemAgent}~\parencite{memagent_2025}, and \textbf{Mem-$\alpha$}~\parencite{mem_alpha_2025}—still treat procedural memory as a static text repository. Their reliance on basic semantic retrieval exposes a critical conceptual flaw: treating retrieved memories as plug-and-play modules rather than raw materials that require adaptation. In reality, procedural memory originates from specific source contexts (e.g., tool configurations and task domains) that often clash with the target environment. By merely injecting static text into the prompt, these methods delegate the complex burden of adaptation to the agent's implicit reasoning. This exposes a critical gap between human intuition and model execution. When a human developer reads a retrieved skill, they naturally mentally adapt its logic to fit a new scenario. We mistakenly assume the LLM will do the same. However, expecting a text prompt to act as an adaptable policy is structurally flawed. In traditional machine learning, policy adaptation~\parencite{lamer_2026} works because parameters are continuous and can be mathematically optimized to fit new environments. In contrast, prompt injection forces the agent to rely on a rigid, discrete text space. Because we cannot continuously optimize static text to align with a dynamic execution state, we have no real guarantee that the skill will generalize safely. Ultimately, what is semantically readable to humans does not equal executable generalization for the model.

To bridge this generalization gap, we must fundamentally shift our conceptualization of procedural memory. Rather than treating the Skill Bank as a static text database, we argue it should be viewed as a repository of encoded behavioral policies, each shaped by its original environment. Once a retrieved skill is understood as an external policy, the challenge of experience reuse naturally transforms into a rigorous problem of \textbf{\textit{policy adaptation}}. In long-horizon tasks like software engineering—which span hours of planning, implementation, testing, and repair—the execution context is inherently dynamic. Code states, local goals, and validation feedback continuously evolve. A policy that provides crucial guidance during the initial scaffolding phase might become actively destructive if applied during late-stage debugging. Therefore, effective experience reuse requires dynamically adapting these encoded policies to fit both specific project configurations and the shifting execution lifecycle. \textbf{Ultimately, long-horizon engineering execution calls for a paradigm shift: from static, rigidly injected skills to dynamically managed, lifecycle-aware guides.}

Motivated by this insight, we introduce \textbf{From Memory to Guide}—a paradigm shift that fundamentally decouples long-term knowledge storage from stage-specific action guidance. To power this transition, we propose the \textbf{Spatio-Temporal Composer for Procedural Coding Memory (Composer)}. Instead of acting as a passive retrieval pipeline, Composer serves as an active policy compiler, transforming rigid, static skills into dynamic, context-aware execution strategies. This dynamic transformation operates across two complementary dimensions:
\begin{itemize}
\item \textbf{Spatial Adaptation (The Environmental Adapter):} Composer aggressively tailors generic skills to the specific configuration of the target workspace, guaranteeing that historical knowledge never clashes with local constraints.
\item \textbf{Temporal Adaptation (The Lifecycle Orchestrator):} Composer synchronizes guidance with the evolving execution phase, precisely dictating not only \textit{when} a skill is triggered, but crucially, \textit{when} it must expire.
\end{itemize}
Synthesizing these dimensions, Composer compiles the raw experience into a \textbf{\textit{Runtime Guide}}—a highly constrained action space equipped with local goals and strict behavioral boundaries. The theoretical implication is profound: Composer morphs static memory into dynamically shifting action trajectories. This achieves true inference-time policy adaptation, unlocking the agent's generalization capabilities without updating a single parameter.

We rigorously evaluate Composer on EngramBench~\parencite{engrambench_2026}, encompassing over 335 hours of execution across 13 demanding, long-horizon software engineering tasks (averaging 4.3 hours per run). Compared to a strong baseline---native Codex equipped with static Skills---Composer achieves a consistently higher confirmed-pass rate across all tasks. Most notably, on the most complex subset of tasks (averaging 8.5 hours per run), it drives an absolute pass-rate increase of 7.2 percentage points while simultaneously slashing the main agent's cumulative token consumption by 32.2\%. Finally, ablation studies empirically validate our dual-axis design: dynamically incorporating \textit{spatial} execution evidence and explicit \textit{temporal} lifecycle boundaries independently yield pass-rate gains of 7.03 and 10.09 percentage points, confirming that our policy adaptation mechanism is both highly effective and architecturally necessary.

Our main contributions are as follows:
\begin{enumerate}
    \item \textbf{We introduce a novel paradigm for procedural memory integration: ``From Memory to Guide''.}
    By rethinking procedural memory not as a static text repository but as an adaptable policy resource, this paradigm fundamentally decouples long-term knowledge storage from stage-specific action guidance.

    \item \textbf{We propose the Spatio-Temporal Composer, an active policy compiler that transforms static memory into dynamic execution guidance.}
    To overcome the rigidity of standard prompt injection, Composer evaluates both environmental constraints (spatial adaptation) and current execution stages (temporal orchestration). It dynamically compiles retrieved experiences into localized \textit{Runtime Guides}, achieving precise, context-aware policy adaptation without a single parameter update.

    \item \textbf{We demonstrate significant gains in both generalization and execution efficiency on demanding long-horizon benchmarks.} 
    Extensive evaluations on 13 complex software engineering tasks in EngramBench reveal that Composer drives an absolute pass-rate increase of 7.2 percentage points on the most challenging subsets. Simultaneously, it slashes the main agent's cumulative token consumption by 32.2\%. Comprehensive ablations further validate the critical necessity and independent performance gains of our dual-axis spatial and temporal design.
\end{enumerate}


\section{Related Work}
\label{sec:related_work}

\paragraph{Advising frozen executors.}
How to Train Your Advisor introduces trainable advisor models that generate instance-specific natural-language guidance for frozen black-box executors~\parencite{advisor_models}.
The advisor is optimized with GRPO using rewards derived from executor outcomes.
Its multi-turn software engineering setting feeds executor actions and observations back to the advisor, and the paper also explores executor-initiated advice requests through tool calls.
Both approaches use generated guidance to adapt execution without updating the executor.
Composer organizes this interface around an explicit separation between persistent procedural knowledge and stage-specific guidance: selected Skills are composed around the current local objective, execution evidence, and previous guidance, while termination and requirement re-entry define the guidance's temporal scope.

\paragraph{Utility-optimized experience generation.}
SEAM generates structured experience for a frozen executor, comprising problem analysis, experience highlights, and a reference plan~\parencite{seam}.
It encodes executor-specific experience in a lightweight generator and optimizes guidance through executor rollouts and GRPO, with optional deployment-time supervised fine-tuning on logged successful
experience.
SEAM and Composer differ in how they organize experience: SEAM internalizes its experience library in generator parameters, whereas Composer retains an explicit Skill Bank as a stable external policy resource.
Composer instantiates selected knowledge as a Runtime Guide whose actions, behavioral boundaries, and termination conditions are tied to a local stage of software development.

\paragraph{Procedural memory and runtime policy adaptation.}
ExpeL extracts reusable insights from agent experience~\parencite{expel}, while Agent Workflow Memory induces workflows and selectively provides them during subsequent interaction in both offline and online settings~\parencite{agent_workflow_memory}.
These approaches establish experience abstraction and reuse as important components of agent adaptation.
From Memory to Guide connects procedural memory with guidance-based adaptation of frozen executors by making the transformation from selected policy knowledge to state-conditioned action guidance explicit.
Its central contribution is the joint organization of guidance content and temporal scope: which experience should inform the current objective, how it should be translated into actions, and when its
local guiding role should end.

\paragraph{Runtime skill interfaces and stage-aware execution.}
SWE-agent demonstrates the importance of agent-computer interfaces for software engineering~\parencite{swe_agent}. SkVM uses model--harness capability profiles to compile Skills and adapts their execution at runtime~\parencite{skvm}, while SkillSmith compiles Skill packages into boundary-guided executable interfaces~\parencite{skillsmith}. Hierarchical Memory Tree organizes reusable web-task knowledge into goals, stages with pre- and post-conditions, and action patterns~\parencite{hierarchical_memory_tree}. Composer focuses on state-conditioned natural-language policy instantiation from a persistent Skill Bank, with explicit local goals, behavioral boundaries, and guidance lifecycles.

\paragraph{Learning experience utility and evolving contexts.}
Memento formalizes memory-augmented decision making with a case-selection policy~\parencite{memento}, while MemRL uses environmental feedback to identify high-utility memories~\parencite{memrl}. Agentic Context Engineering evolves contextual playbooks through generation, reflection, and curation~\parencite{agentic_context_engineering}. These approaches develop the selection and organization of experience; Composer makes the subsequent transformation of selected knowledge into local, time-bounded action guidance explicit.

\section{From Memory to Guide: The Spatio-Temporal Composer}
\label{sec:memory_to_guide}

We introduce a new paradigm that restructures procedural memory from static experience delivery to context-conditioned policy adaptation. This section formally defines this paradigm shift and details the architecture of our proposed system.

\subsection{Problem Formulation: Rethinking Procedural Memory}
\label{sec:problem_formulation}

Consider a task $T$ with profile $p_T$ (objectives, rules, tool conditions). At step $t$, the execution agent generates an action $a_t$ based on the dynamic interaction context $h_t$. Let $\pi_\theta$ denote the base execution policy and $\mathcal{K}=\{s_1,s_2,\ldots,s_N\}$ be the procedural memory bank, where each Skill $s_i$ encodes specific operational experience.

Under native Skill use, the Skill Bank acts merely as a static text repository retrieved on demand. The execution policy is expressed as:
\begin{equation}
a_t \sim \pi_{\theta,\mathcal{K}}^{\mathrm{native}} (\cdot\mid p_T,h_t) = \pi_\theta \bigl( \cdot\mid p_T,h_t, \mathcal{D}(\mathcal{K},p_T,h_t) \bigr),
\tag{1}
\label{eq:native_skill_policy}
\end{equation}
where $\mathcal{D}$ is the retrieval mechanism. This formulation forces the base model $\pi_\theta$ to implicitly interpret and adapt the retrieved knowledge $\mathcal{D}(\cdot)$ at every step, frequently leading to context mismatches and negative transfer in long-horizon tasks.

To solve this, we explicitize the interpretation process through runtime adaptation. At request step $\tau$, given execution evidence $o_\tau$ extracted from $h_\tau$ and the previous Guide $g_{\mathrm{prev}}$, we formulate adaptation as:
\begin{equation}
g_\tau = \mathcal{C}_\phi \left( p_T,\mathcal{K},o_\tau,g_{\mathrm{prev}} \right), \quad g_\tau \in \mathcal{G}\cup\{\varnothing\},
\tag{2}
\label{eq:runtime_adaptation}
\end{equation}
where $\mathcal{C}_\phi$ is the runtime adapter and $\varnothing$ indicates no new guidance is needed. Each generated Guide is a structured policy interface:
\begin{equation}
g_\tau= \left( \ell_\tau, r_\tau, b_\tau, e_\tau \right),
\tag{3}
\label{eq:guide_structure}
\end{equation}
where $\ell_\tau$ defines the local goal, $r_\tau$ provides state-conditioned recommended actions, $b_\tau$ specifies behavioral boundaries and applicability prerequisites, and $e_\tau$ defines the termination conditions. Let $g_t^{\mathrm{act}}$ denote the guidance currently active at step $t$. The execution policy under our paradigm becomes:
\begin{equation}
a_t \sim \pi_\theta \left( \cdot\mid p_T,h_t,g_t^{\mathrm{act}} \right).
\tag{4}
\label{eq:guided_execution_policy}
\end{equation}

By decoupling adaptation ($\mathcal{C}_\phi$) from execution ($\pi_\theta$), this paradigm transforms implicit experience delivery into explicit, stage-specific execution. To realize this shift, we must solve the \textbf{spatio-temporal adaptation problem}: matching the content of experience to the target environment's spatial constraints (project rules, interfaces), and managing its lifecycle along the temporal execution timeline (local goals, evolving code states).

\subsection{Spatio-Temporal Composer Architecture}
\label{sec:composer_architecture}

To realize the runtime adapter $\mathcal{C}_\phi$ defined above, we propose the Spatio-Temporal Composer for Procedural Coding Memory (Composer). Composer acts as an explicit interface that transforms static skills into dynamic action guidance, separating the selection of reusable knowledge from its instantiation as state-conditioned actions.

\begin{figure*}[t]
\centering
\includegraphics[width=\textwidth]{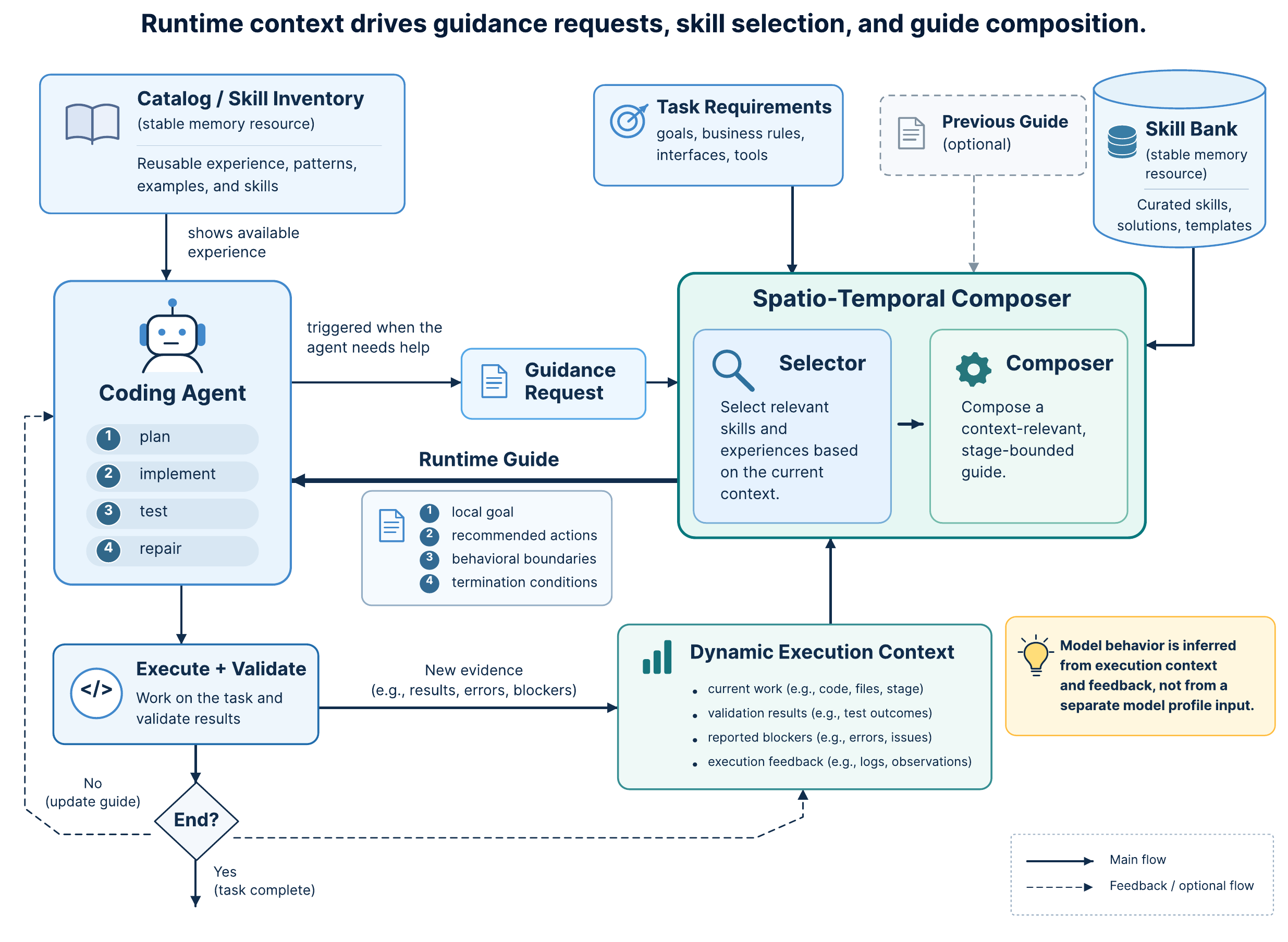}
\caption{
Overview of Spatio-Temporal Composer.
The Catalog exposes available capabilities to the Coding Agent, which initiates guidance requests during development.
Task requirements, dynamic execution context, and the previous Guide inform skill selection and guide composition.
The resulting Runtime Guide is returned to the Coding Agent, whose subsequent execution and validation provide evidence for future requests.
}
\label{fig:composer_framework}
\end{figure*}

As illustrated in Figure~\ref{fig:composer_framework}, a compact Catalog exposes the reusable capabilities available in $\mathcal{K}$. The Coding Agent synchronously initiates guidance requests based on its local goal, progress, and blockers. At request step $\tau$, Composer synthesizes three complementary sources of information:
\begin{enumerate}
\item \textbf{Task profile ($p_T$)}: Project objectives, business rules, interface constraints, and tool conditions.
\item \textbf{Execution evidence ($o_\tau$)}: The agent's description of current work, session context, reported validation results, and blockers.
\item \textbf{Previous Guide ($g_{\mathrm{prev}}$)}: The earlier local goal and recommendations, allowing Composer to assess how previous guidance relates to completed work and current gaps.
\end{enumerate}

\subsection{Context-Conditioned Skill Selection}
\label{sec:skill_selection}

The Selector turns a guidance request into a concrete composition directive. It evaluates the available Skills in $\mathcal{K}$ against $p_T$, $o_\tau$, and $g_{\mathrm{prev}}$, returning three structured fields:

\noindent\textbf{Guidance decision (\texttt{decision}):} Specifies whether to generate a new Guide. The Selector considers current goals against outcomes reported for earlier guidance. If new guidance is warranted, the request proceeds; otherwise, existing guidance remains governed by its own termination conditions.

\noindent\textbf{Skill references (\texttt{skillRefs}):} Identifies the specific Skills in $\mathcal{K}$ to use for the current request. Multiple Skills can be selected when their rules contribute complementary components of the required guidance.

\noindent\textbf{Rationale-preserving handoff (\texttt{whyNow}):} Records the exact reason for the guidance decision, identifying the current execution gap and the intended contribution of the newly selected Skills relative to the previous Guide.

This structure strictly separates \emph{which} experience to use (\texttt{skillRefs}) from \emph{why} it is needed now (\texttt{whyNow}), bridging selection and composition without information loss.

\subsection{Spatial Composition and Temporal Adaptation}
\label{sec:spatial_temporal_adaptation}

Composer translates the selected Skills into the final Runtime Guide ($g_\tau$) by addressing both the spatial context and the temporal lifecycle.

\noindent\textbf{Spatial Composition: From Rules to Situated Actions.}
Composer instantiates general rules as situated actions around the local goal $\ell_\tau$. Driven by the \texttt{whyNow} rationale, spatial composition maps abstract experience to concrete project entities. It aligns related rules into coherent recommended actions ($r_\tau$) while enforcing behavioral boundaries ($b_\tau$) dictated by the task profile $p_T$.

\begin{figure*}[t]
\centering
\includegraphics[width=0.96\textwidth]{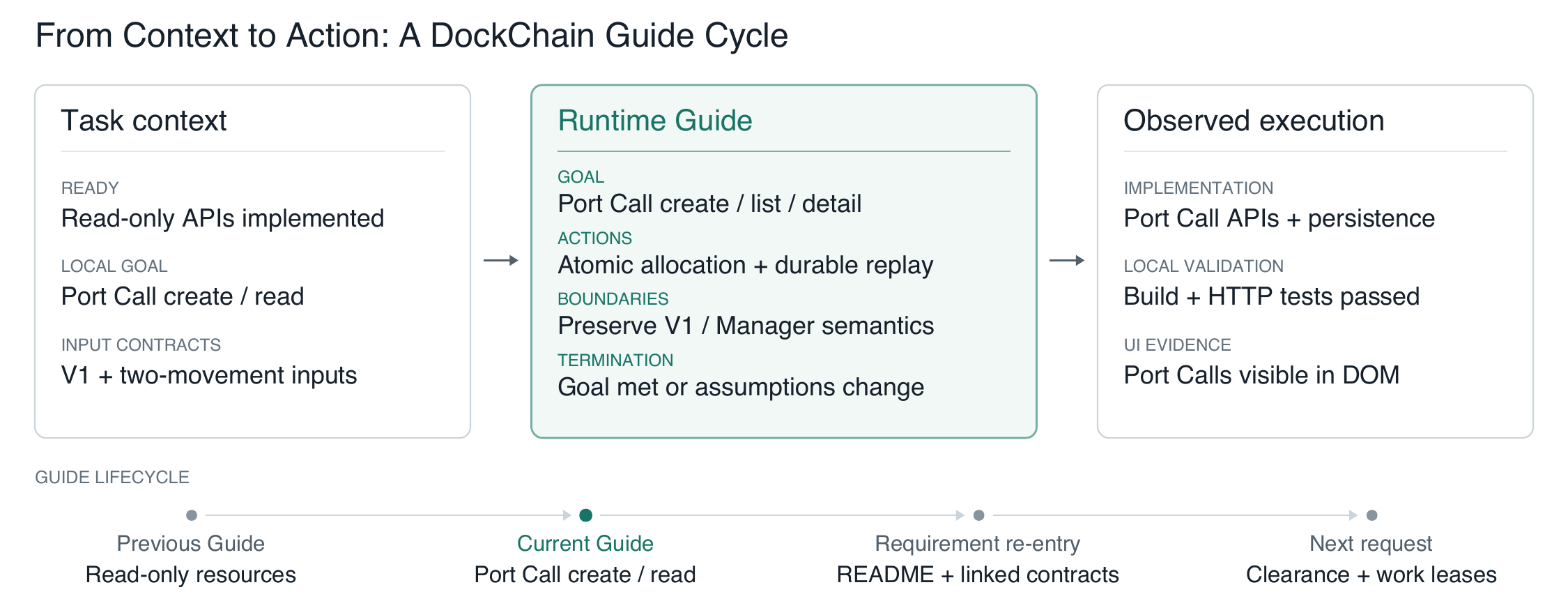}
\caption{
A DockChain example of spatial composition and temporal adaptation:
execution context is instantiated as a local Runtime Guide, whose completion
triggers requirement re-entry and the next guidance request.
}
\label{fig:dockchain_guide_cycle}
\end{figure*}

The upper portion of Figure~\ref{fig:dockchain_guide_cycle} illustrates this in the DockChain task. When the goal shifts to Port Call creation, the Guide translates general consistency and idempotency skills into a specific PostgreSQL transaction allocating resources, writing states, and persisting responses. It explicitly dictates interface semantics for both V1 and two-movement inputs, preventing spatial context mismatch (e.g., executing incompatible database commands).

\noindent\textbf{Temporal Adaptation: Guide Lifecycle Management.}
To prevent out-of-phase guidance from disrupting execution, the temporal dimension bounds the Runtime Guide with explicit lifecycle specifications ($e_\tau$ and the applicability prerequisites in $b_\tau$).

The execution agent uses \texttt{Applies When}, \texttt{Verification}, and \texttt{Termination} clauses to independently determine whether guidance should continue. Goal completion, changes in the execution stage, or insurmountable blockers trigger reassessment. As shown in the lower part of Figure~\ref{fig:dockchain_guide_cycle}, once the Port Call implementation and validation are completed, the local guidance terminates. The agent rereads the task requirements, identifies the next local goal (clearance and lease handling), and issues a new request accompanied by the latest execution evidence.

\begin{algorithm}[t]
\caption{Runtime Procedure of Spatio-Temporal Composer}
\label{alg:composer}
\small
\begin{algorithmic}[1]
\Require Task profile $p_T$, Skill Bank $\mathcal{K}$, initial context $h_0$,
agent $A$, Selector $S$, Composer $C$
\State $t \gets 0,\quad g_0^{\mathrm{act}} \gets \varnothing,\quad g_{\mathrm{prev}} \gets \varnothing$
\State Provide $A$ with $\operatorname{Catalog}(\mathcal{K})$

\While{task execution continues}
\State $A$ identifies the current local goal from $h_t$
\If{$A$ requests guidance}
\State $\tau \gets t$
\State $o_\tau \gets \operatorname{Evidence}(h_\tau)$
\State $(d_\tau,R_\tau,w_\tau) \gets S(p_T,\mathcal{K},o_\tau,g_{\mathrm{prev}})$
\If{$d_\tau=\textsc{Generate}$}
\State $g_\tau \gets C(p_T,\mathcal{K}[R_\tau],o_\tau,g_{\mathrm{prev}},w_\tau)$
\State $g_t^{\mathrm{act}} \gets g_\tau$
\Else
\State $g_\tau \gets \varnothing$
\EndIf
\EndIf
\State $h_{t+1} \gets A.\operatorname{Execute}(p_T,h_t,g_t^{\mathrm{act}})$
\State $g_{t+1}^{\mathrm{act}} \gets g_t^{\mathrm{act}}$
\If{$g_t^{\mathrm{act}} \neq \varnothing$ and $g_t^{\mathrm{act}}$ terminates}
\State $g_{\mathrm{prev}} \gets g_t^{\mathrm{act}}$; $g_{t+1}^{\mathrm{act}} \gets \varnothing$
\State Perform requirement re-entry and update $h_{t+1}$
\EndIf
\State $t \gets t+1$
\EndWhile
\end{algorithmic}
\end{algorithm}

In Algorithm~\ref{alg:composer}, $t$ indexes execution steps, and $\tau=t$ records the step at which a guidance request occurs; request times therefore form a subset of execution steps. The request-specific output $g_\tau$ is distinct from the active Guide $g_t^{\mathrm{act}}$: when no new Guide is generated, existing guidance remains active until its termination conditions are met. This continuous loop of context-conditioned generation, bounded execution, and requirement re-entry ensures that procedural memory acts not as a static constraint, but as an evolving policy perfectly synchronized with the task's spatial requirements and temporal progress.
\section{Experimental Setup}
\label{sec:experimental_setup}

We evaluate whether runtime policy adaptation improves the use of a fixed Skill Bank in long-horizon software engineering, and examine the contributions of contextualized composition, dynamic execution evidence, and guidance lifecycles.

\subsection{Tasks and Execution Protocol}
\label{sec:tasks_protocol}

We use all 13 transfer tasks in EngramBench~\parencite{engrambench_2026}, covering diverse business domains, interface contracts, and state-transition requirements.
Each task requires sustained implementation, testing, and repair against its project requirements.
The execution agent is Codex with GPT-5.5 at \texttt{medium} reasoning effort.
The Skill Bank contains 18 reusable engineering Skills, constructed before transfer evaluation and held fixed throughout execution.

The main comparison uses the same execution model and Skill Bank, with three runs per task and condition, totaling 78 development runs.
Final project artifacts are assessed by task-specific evaluators.
Runtime guidance uses development-time requirements and execution evidence; final hidden-test outcomes are reserved for evaluation.


\subsection{Compared Methods}
\label{sec:compared_methods}

\paragraph{Main comparison.}
\textbf{Codex + Skills} uses the native Skill mechanism to access the fixed bank, with the execution agent interpreting and applying the supplied experience.
\textbf{Composer} uses the same bank through runtime selection and composition, delivering state-conditioned Runtime Guides.

\paragraph{Component controls.}
\textbf{Selector-only} selects and directly delivers Skill content without Guide composition, testing the contribution of contextualized composition beyond selection.
\textbf{Generic Advisor} generates runtime advice without Skill input, testing the contribution of Skill-grounded guidance beyond generic advice.

\paragraph{Design ablations.}
\textbf{w/o Dynamic Evidence} retains the Selector, Skill Bank, and auxiliary-model configuration while removing the composition module's direct access to current execution progress and feedback.
It tests the contribution of dynamic evidence to state-conditioned composition.
\textbf{w/o Lifecycle} retains dynamic Guide generation but removes explicit local termination and requirement re-entry specifications, testing the contribution of guidance lifecycles.
Global task requirements remain in effect.

Component controls and design ablations use DockChain, CarbonLedger, EscrowGuard, and FlagFoundry, with three runs per task and condition.
The design ablations retain the other components of the full Composer configuration.

\subsection{Evaluation Metrics}
\label{sec:evaluation_metrics}

\paragraph{Task quality.}
We report each task's mean confirmed-pass rate across three runs. The rate is the number of confirmed \texttt{passed} outcomes divided by all scheduled test executions, including \texttt{excluded} and diagnostic instances. Excluded executions remain in the denominator without being classified as observed implementation failures. Aggregate rates pool counts across the stated tasks and runs under the same rule for all conditions; we distinguish relative count improvements from percentage-point changes in rate.

\paragraph{Execution-side token efficiency.}
We measure the main Coding Agent's cumulative input and output tokens across development.
This execution-side metric quantifies the token usage associated with the submitted implementations and is distinct from auxiliary-model orchestration costs.

\paragraph{Development time.}
We report per-run development duration and cumulative time across runs, including tool execution and Guide waiting.
Final hidden-test evaluation is excluded.
Cumulative time is computed by summing individual run durations and characterizes the execution horizon independently of experiment parallelism.


\section{Results and Analysis}
\label{sec:results}

\subsection{Task Quality and Execution-Side Efficiency}
\label{sec:overall_results}

Composer consistently outperforms the native Codex + Skills baseline across all 13 demanding EngramBench tasks, validating the robustness of runtime policy adaptation.
The main comparison comprises 78 development runs, totaling more than 335 cumulative development run-hours, averaging over 4.3 hours per run, with the longest run exceeding 12 hours.

\begin{figure*}[t]
    \centering
    \includegraphics[width=\textwidth]
        {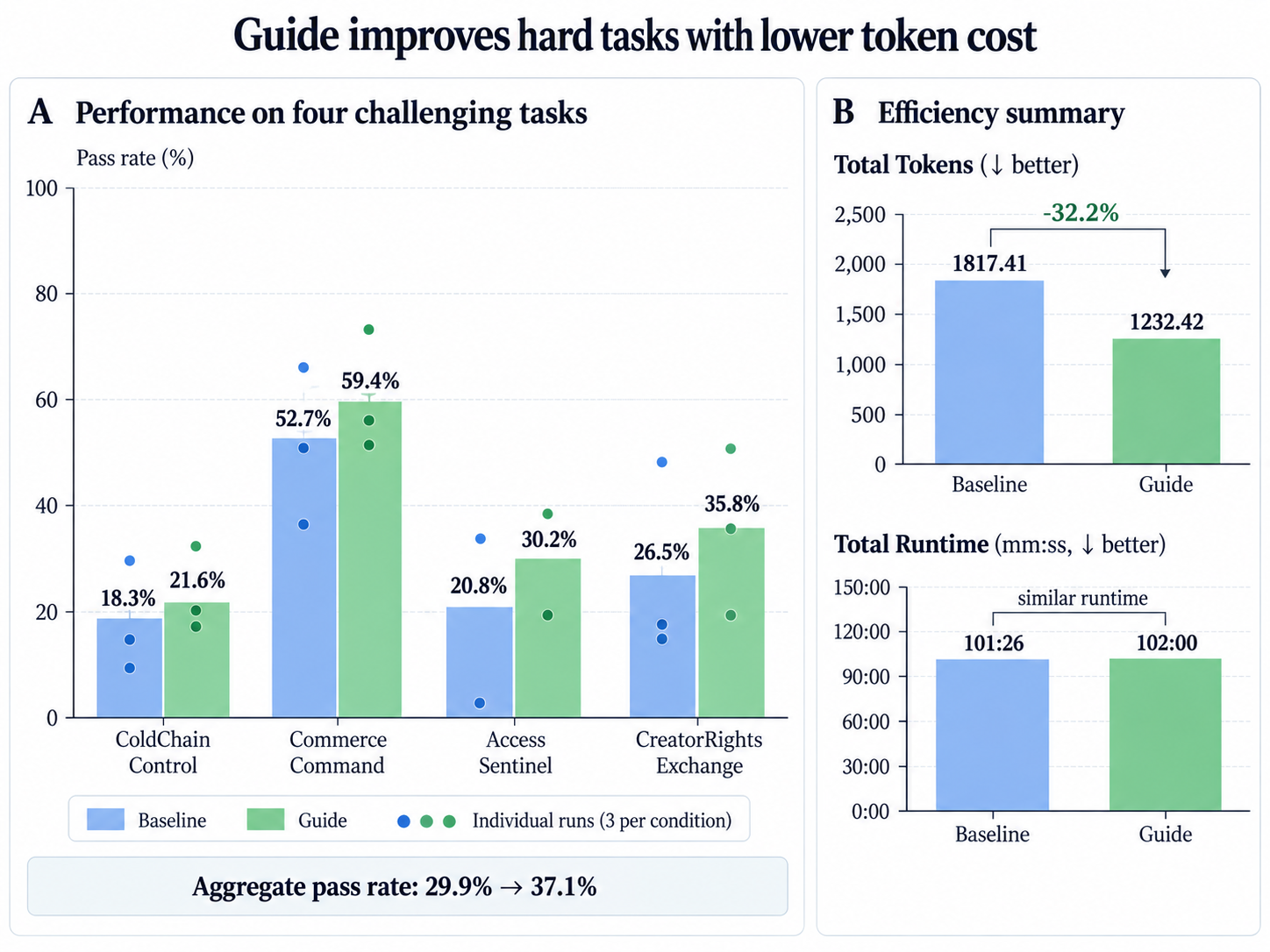}
    \caption{
        Task quality and execution-side efficiency on four tasks
        from the 13-task EngramBench evaluation.
        Left: mean confirmed-pass rates across three runs per condition,
        with individual runs shown as dots.
        Right: cumulative Coding Agent token usage and development
        runtime (hh:mm) across 12 runs per condition.
        Baseline denotes Codex + Skills; Guide denotes Composer.
    }
    \label{fig:quality_efficiency}
\end{figure*}

Figure~\ref{fig:quality_efficiency} presents four tasks within this benchmark whose development runs average 8.48 hours.
Composer increases the aggregate confirmed-pass rate from 29.9\% to 37.1\%, a gain of \textbf{7.2 percentage points}, while reducing the main Coding Agent's cumulative token usage by \textbf{32.2\%}.
Improvements occur on every task shown, including gains of 9.4 percentage points on AccessSentinel and 9.3 percentage points on CreatorRightsExchange.
Composer therefore delivers higher task quality with fewer execution-side tokens over extended development workflows.

The same aggregate quality--efficiency pattern holds across the remaining nine tasks.
Cumulative confirmed passes increase from 458 to 563, a relative improvement of \textbf{22.9\%}, while main-agent token usage decreases by \textbf{6.86\%}.

\subsection{Beyond Skill Selection and Generic Advice}
\label{sec:component_results}

Table~\ref{tab:component_ablation_results} compares Composer with native Codex + Skills, Selector-only, and Generic Advisor on four tasks, with three runs per task and condition.

Using the same Skill Bank as native Codex + Skills, Composer increases cumulative confirmed passes from 173 to 229, a relative gain of \textbf{32.37\%}.
Selector-only records 149 confirmed passes, while Generic Advisor records 157.
Composer therefore adds 80 confirmed passes over Selector-only and 72 over Generic Advisor, outperforming all three component-comparison baselines on each task.
The gains over both controls demonstrate the incremental value of contextualized composition: selected policy knowledge becomes more effective when organized into guidance for current execution.

\begin{table*}[t]
    \centering
    \small
    \setlength{\tabcolsep}{5pt}
    \renewcommand{\arraystretch}{1.12}
    \caption{
        Component comparisons and design ablations.
        Task rows report cumulative confirmed passes across three
        runs per condition; parentheses indicate scheduled tests per run.
        Each condition comprises 555 scheduled test executions.
        Bold denotes the highest value in each row.
        ``w/o Direct Evidence'' removes direct dynamic execution
        evidence from composition; ``w/o Lifecycle'' removes explicit
        termination and requirement re-entry specifications.
    }
    \label{tab:component_ablation_results}
    \begin{tabular}{@{}lrrrrrr@{}}
        \toprule
        Task
        & \shortstack{Codex +\\Skills}
        & \shortstack{Selector-\\only}
        & \shortstack{Generic\\Advisor}
        & \shortstack{w/o Direct\\Evidence}
        & \shortstack{w/o\\Lifecycle}
        & Composer \\
        \midrule
        DockChain (44)
            & 64 & 33 & 49 & 65 & 55 & \textbf{69} \\
        CarbonLedger (49)
            & 49 & 75 & 62 & \textbf{81} & 67 & 80 \\
        EscrowGuard (48)
            & 21 & 7 & 3 & 9 & 23 & \textbf{32} \\
        FlagFoundry (44)
            & 39 & 34 & 43 & 35 & 28 & \textbf{48} \\
        \midrule
        Total confirmed passes
            & 173 & 149 & 157 & 190 & 173 & \textbf{229} \\
        Confirmed-pass rate (\%)
            & 31.17 & 26.85 & 28.29 & 34.23 & 31.17
            & \textbf{41.26} \\
        \bottomrule
    \end{tabular}
\end{table*}

\subsection{Contributions of Execution Evidence and Lifecycle}
\label{sec:design_ablation_results}

The full Composer achieves the highest aggregate confirmed-pass rate in Table~\ref{tab:component_ablation_results}, outperforming both design ablations.

\paragraph{Dynamic execution evidence.}
Providing the composition module with direct dynamic execution evidence increases cumulative confirmed passes from 190 to 229.
The aggregate confirmed-pass rate rises from 34.23\% to 41.26\%, a gain of \textbf{7.03 percentage points}.
The improvement occurs on three of the four tasks.
This aggregate gain demonstrates the value of conditioning Guide composition on current progress and feedback beyond Skill selection.

\paragraph{Explicit guidance lifecycle.}
Retaining explicit termination and requirement re-entry specifications increases cumulative confirmed passes from 173 to 229.
The aggregate confirmed-pass rate rises from 31.17\% to 41.26\%, a gain of \textbf{10.09 percentage points}, with higher counts on all four tasks.
Because this ablation retains dynamic Guide generation, the comparison demonstrates the contribution of lifecycle specifications beyond generating state-conditioned advice.

Dynamic evidence improves how policy knowledge is organized for current execution, while lifecycle specifications connect local guidance to stage transitions and requirement reassessment.

\subsection{Guides in Action}
\label{sec:trajectory_analysis}

Figure~\ref{fig:dockchain_guide_cycle} connects these mechanisms to an observed DockChain execution.
The Port Call Guide binds transaction consistency, idempotency, and interface compatibility to the current local objective.

The Guide is followed by Port Call implementation and local validation through tests and UI checks.
After reporting its local implementation and validation results, the agent rereads the README and referenced requirements, then submits a new guidance request for clearance and lease handling.

Reusable requirements and engineering principles persist across development, while each Guide organizes the actions appropriate to the current stage.

\section{Toward Self-Evolving Procedural Policies}
\label{sec:future_work}

Runtime Guides make it easier for the system to learn from its own actions because each Guide clearly defines a goal, boundaries, and when to stop. In future work, we can use these well-defined steps to test different ways of applying the same skill and compare the results side-by-side.

To make this work, we can borrow ideas from recent research on memory credit assignment~\parencite{memq} and text-based feedback~\parencite{protegi, textgrad}. In simple terms, we can take successful task outcomes and text feedback and turn them into training data. We can then use this data to fine-tune the Composer. This means the Composer itself can be continuously improved and optimized, even while the Skill Bank and the main agent remain completely fixed.

This setup paves the way for true recursive self-improvement (RSI). The system can form a complete learning loop: it generates a guide, tests it in the environment, and uses the feedback to update the Composer. This allows the system to issue better guides over time, automatically adjusting how strict or detailed it needs to be as the main agent's abilities improve. Ultimately, our goal is to build a self-evolving memory system---one that not only stores useful skills but constantly learns the best way to apply them.

\section{Conclusion}
\label{sec:conclusion}

In this work, we introduced \textit{From Memory to Guide}, a new paradigm that fundamentally separates persistent procedural knowledge from context-specific action guidance. We realized this through the Spatio-Temporal Composer, which treats the Skill Bank not as a static text repository, but as an adaptable policy resource. By actively aligning past experiences with current environmental constraints (spatial adaptation) and strictly defining when that guidance applies (temporal adaptation), Composer dynamically compiles raw skills into bounded \textit{Runtime Guides}. Together, these mechanisms achieve true inference-time policy adaptation without requiring any updates to the execution model's parameters.

Extensive evaluations on demanding, long-horizon software engineering tasks in EngramBench demonstrate the clear advantages of this architecture. Composer consistently outperformed the native Codex + Skills baseline across all evaluated tasks. Most notably, on the most complex subset of tasks, it increased the aggregate confirmed-pass rate by 7.2 percentage points while simultaneously slashing the main agent's token usage by 32.2\%. Comprehensive ablations further confirmed that both spatial grounding and temporal lifecycles are indispensable for reliable execution. 

Ultimately, \textit{From Memory to Guide} advances procedural memory from passive experience delivery to active runtime policy adaptation: Memory preserves reusable policy knowledge, while Guide gives that knowledge its actionable form and stage-specific scope.

\printbibliography[heading=bibnumbered,title={References}]

\end{document}